\def\year{2022}\relax
\documentclass[letterpaper]{article} 
\usepackage{aaai22}  
\usepackage{times}  
\usepackage{helvet}  
\usepackage{courier}  
\usepackage[hyphens]{url}  
\usepackage{graphicx} 
\usepackage{natbib}  
\usepackage{caption} 
\DeclareCaptionStyle{ruled}{labelfont=normalfont,labelsep=colon,strut=off} 
\usepackage{epsfig}
\usepackage{graphicx}
\usepackage{amsmath}
\usepackage{amssymb}
\usepackage{siunitx, mhchem}
\usepackage{array}\usepackage{makecell}
\usepackage{subfig}

\newcommand{\com}[1]{}

\usepackage{algorithm}
\usepackage{algorithmic}

\usepackage{newfloat}
\usepackage{listings}
\floatstyle{ruled}
\newfloat{listing}{tb}{lst}{}
\floatname{listing}{Listing}
\title{Dodging Attack Using Carefully Crafted Natural Makeup}

\author {
    Nitzan Guetta,\textsuperscript{\rm 1}
    Asaf Shabtai,\textsuperscript{\rm 1}
    Inderjeet Singh,\textsuperscript{\rm 2}
    Satoru Momiyama,\textsuperscript{\rm 2}
    Yuval Elovici\textsuperscript{\rm 1}
}
\affiliations {
    \textsuperscript{\rm 1} Ben-Gurion University of the Negev\\
    \textsuperscript{\rm 2} NEC Corporation\\
}

\begin{document}

\maketitle

\begin{abstract}
Deep learning face recognition models are used by state-of-the-art surveillance systems to identify individuals passing through public areas (e.g., airports). 
Previous studies have demonstrated the use of adversarial machine learning (AML) attacks to successfully evade identification by such systems, both in the digital and physical domains.
Attacks in the physical domain, however, require significant manipulation to the human participant's face, which can raise suspicion by human observers (e.g. airport security officers).
In this study, we present a novel black-box AML attack which carefully crafts natural makeup, which, when applied on a human participant, prevents the participant from being identified by facial recognition models. 
We evaluated our proposed attack against the ArcFace face recognition model, with 20 participants in a real-world setup that includes two cameras, different shooting angles, and different lighting conditions.
The evaluation results show that in the digital domain, the face recognition system was unable to identify all of the participants, while in the physical domain, the face recognition system was able to identify the participants in only 1.22\% of the frames (compared to 47.57\% without makeup and 33.73\% with random natural makeup), which is below a reasonable threshold of a realistic operational environment.

\end{abstract}

\section{Introduction}

\begin{figure}[t]
    \includegraphics[width=0.99\linewidth]{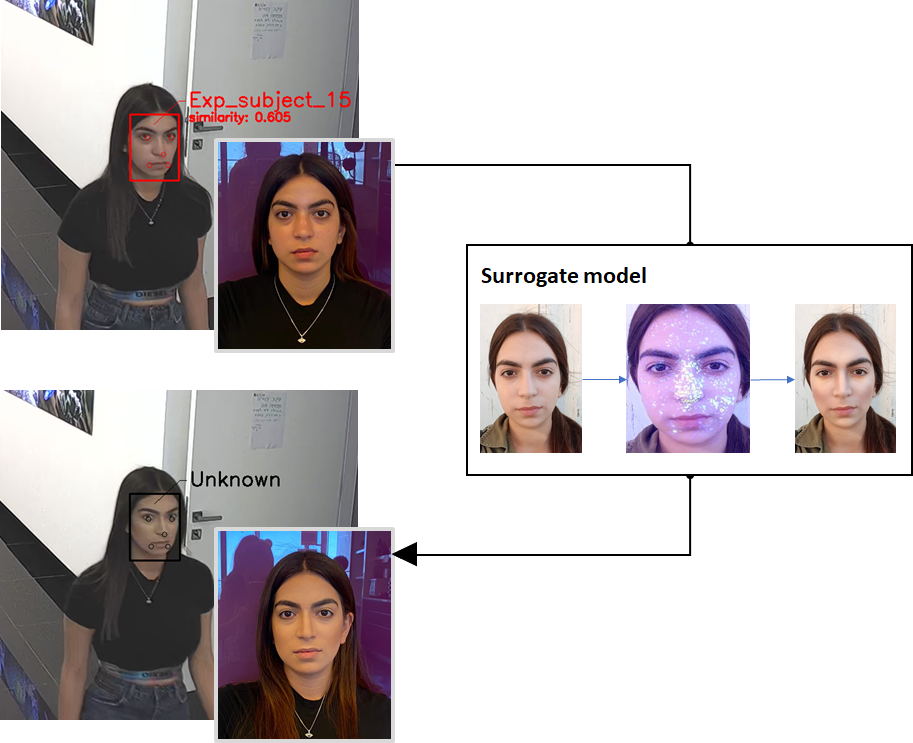}
    \caption{In the upper image the attacker is recognized by the face recognition (FR) system.
    In the middle image, our method uses a surrogate model to calculate the adversarial makeup in the digital domain, that is then applied in the physical domain . As a result, the attacker is not identified by the FR system (lower image).
}
\label{attack demo}
\end{figure}

Face recognition (FR) system are widely used in public areas, including subways, airports, and workplaces, to automatically identify individuals~\cite{security1,security3,security2}.
State-of-the-art FR systems are based on deep neural networks (DNNs), which have demonstrated high accuracy, outperforming traditional face recognition algorithms~\cite{FaceRecognitionNN3,FaceRecognitionNN1}, especially due to recent improvements in the loss functions used in training methods~\cite{L-Softmax,liu2017sphereface,AM-Softmax1,AM-Softmax2,ArcFace} and the large datasets available for training~\cite{lfw,CASIA-WebFace,Ms-celeb-1m}. 

In recent years, DNN models have shown to be vulnerable to adversarial machine learning (AML) attacks~\cite{dnn-vulnerabilities}.
Such attacks have also been demonstrated on face recognition models~\cite{FR-vulnerable-survey}.
Most of the previously presented attacks focused on the digital domain, i.e., manipulating the digital image of an individual~\cite{digital_attack1,digital_attack2,digital_attack3,digital_attack4,digital_attack5, makeup-attack-digital}. 
However, in most cases, the attacks involved adding small perturbations to the image, which is very hard to transfer to the physical domain~\cite{digital_attack2, digital_attack3, digital_attack4,digital_attack5}.

In the limited number of studies demonstrating real-world physical attacks, the attacks were implemented using patches, printed on various wearable objects such as glasses or hats, that contain the attack perturbation~\cite{advmakeup2021,glasses-attack1,makeup-attack-physical,patches-attack1,patches-attack2}.
Such patches may result in a high evasion rate, but they will likely raise suspicion, since they are far from being inconspicuous.

We assume that surveillance systems implemented in places such as airports are secured, and therefore, digital attacks would be more difficult to execute. 
While physical world adversarial attacks are more practical and easier to execute, the attacker must consider the fact that the physical patch might look suspicious and could be detected by an observer (e.g., a human supervisor).   
Moreover, in many cases, a person wearing a hat or glasses is asked to remove them during the identification process. 

In this paper, we propose a dodging adversarial attack that is black-box, untargeted, and based on a perturbation that is implemented using natural makeup.
Since natural makeup is physically inconspicuousness, its use will not raise suspicion.
Our method finds a natural-looking makeup, which, when added to an attacker's face, hides his/her identity from a face recognition system. 
An example of the attack, as captured by a camera in our experimental setup, is presented in Figure~\ref{attack demo}.

The proposed attack consists of the following two main phases: (1) The offline preparation phase: Given an image of the attacker, our method identifies the areas of the face that have the greatest influence on the identification decision by the FR surrogate model. 
Then, digital makeup is added to these areas in order to hide the identity of the attacker from the FR system. 
These two steps are repeated until the surrogate FR model cannot correctly identify the attacker. 
(2) The online execution phase: Based on the image with the digital makeup, makeup is physically applied on the attacker with the assistance of a makeup artist. 
Using the physical makeup, the attacker attempts to evade identification by the target FR model.

The proposed attack was evaluated with 20 participants (an equal number of males and females), using a realistic setup including two cameras located in a corridor, with different shooting angles and lighting conditions.
For the attack's surrogate model, we used the \textit{Facenet} FR model, and ~\textit{LResNet100E-IR,ArcFace@ms1m-refine-v2} served as the target model. 
For each participant the attack was first applied digitally and then physically with the assistance of a makeup artist.

Experimental results show that the proposed method achieves 100\% success in the digital domain on both the surrogate and target models. 
In the physical experiments, on average, the FR system identifies the attackers without any makeup in 47.57\% of the frames in which a face was detected, with random makeup in 33.73\% of the frames, and using our attack in only 1.22\% of the frames, which is below a reasonable threshold of a realistic operational environment.

\noindent We summarize the contributions of our work as follows: 
\begin{itemize}
    \item a novel black-box AML attack that carefully crafts natural makeup, which, when applied on a human participant, results in successfully evading identification by face recognition models;
    \item an AML attack that works in a realistic setup that includes different cameras and shooting angles;
    \item an AML attack in the physical domain that is inconspicuous, unlike attacks presented in previous studies.
\end{itemize}

\section{\label{section:background}Background}
\begin{figure*}[t!]
    \begin{center}
    \includegraphics[width=0.99\linewidth]{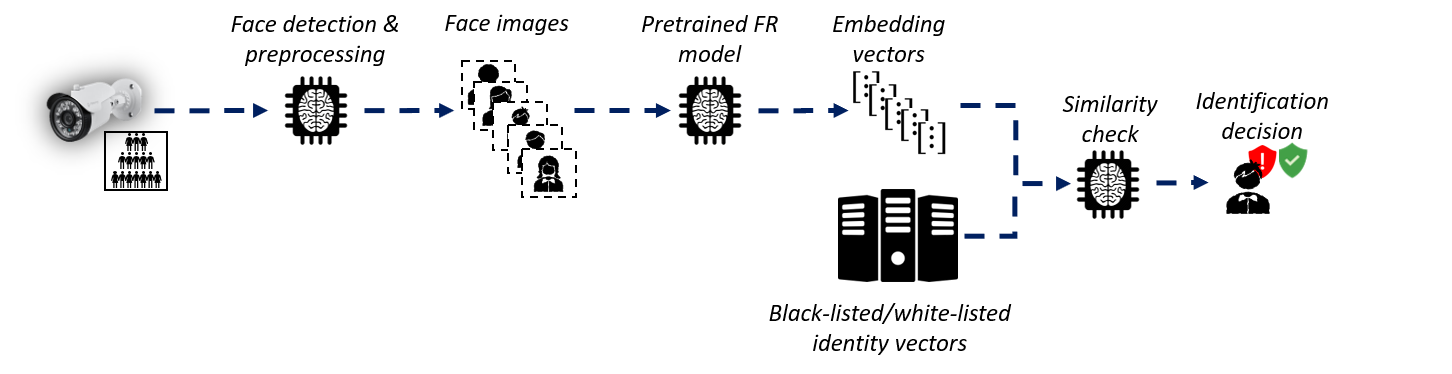}
    \end{center}
    \caption{FR system identification process.}
\label{FR_pipeline}
\end{figure*}

\subsection{Face Recognition Systems}
Typically, FR systems are used to identify white-listed individuals and/or raise alerts about black-listed individuals captured by the system's cameras. Usually, FR systems are made up of several components, as shown in Figure~\ref{FR_pipeline}. The system's input is provided by cameras that produce a video stream which includes a sequence of frames. These frames are passed to a face detection component, which usually is the first step of any fully automated FR system \cite{MTCNN,detection1,detection2,detection3}. 
The face detector produces facial landmarks that are used for pre-process.
The pre-process converts each face in the image individually to an input image that is passed to the FR component for analysis.

The model behind the FR component is usually implemented using a DNN. This model is based on an embedding function $f$ that maps a face image $x$ into a vector $f(x)$. This vector representation allows comparison between different face images. 
In the identification component, the resulting vectors are compared to vectors of black-listed/white-listed individuals, information which is predefined in the FR system. Eventually, the identity of an individual is determined based on the highest similarity score that is passing a certain threshold.

\subsection{Triplet Loss}
The triplet loss function is used to optimize an embedding function $f(x)$. 
This loss function was first proposed by~\citet{FaceRecognitionNN2}, and it operates on three examples from the dataset, referred to as triplets.
An FR model encodes a face image $x$ into a $d$-dimensional space, so the embedding feature is represented by $f(x) \in R^d$. 
Given an embedding function $f(x)$ and three face images: (a) \textbf{anchor $x_i^a$} - an image of a certain individual, (b) \textbf{positive $x_i^p$} - an additional image of this individual, and (c) \textbf{negative $x_i^n$} - an image of another individual, the following equation must be satisfied for all possible triplets:
$$||f(x_i^a) - f(x_i^p)||_2^2 + \alpha \leq ||f(x_i^a) - f(x_i^n)||_2^2$$
where $\alpha$ is a margin between positive and negative pairs.

This leads to the following cost function, which can be used to minimize an optimization problem:
$$\sum_{i}^{N} \max(0,||f(x_i^a) - f(x_i^p)||_2^2 - ||f(x_i^a) - f(x_i^n)||_2^2 + \alpha)$$
The triplet loss function is used to reduce the distance between the anchor and the positive identity; at the same time it is used to increase the distance between the anchor and the negative identity.

\section{Related Work}\label{section:related work}
\subsection{Attacks Against Face Recognition Models}\label{subsection:FR attacks}
AML attacks in the face recognition domain have been widely investigated in the last few years \cite{FR-vulnerable-survey}. These attacks can be divided into two categories. The first category includes studies in which a type of digital manipulation is performed on a digital image, including the addition of some carefully crafted noise to an image. The second category includes studies where physical manipulation of objects in the real world is performed, such as printing special patterns on eyeglass frames and wearing them before taking a picture.

\subsubsection{AML attacks in the digital domain}\label{subsubsection:Digital attacks}      
A recent study \cite{digital_attack1} employed a conditional variation autoencoder generative adversarial network (GAN) to craft targeted adversarial face images in a semi-white-box setting. In another study, \cite{digital_attack2} a GAN was used to generate adversarial face images to evade a FR system. In contrast to the previous study, the face images crafted appear to be more natural and differ very slightly from the target face images. In \cite{digital_attack3}, an evolutionary optimization method was adopted to generate adversarial samples in a black-box setting. Another study introduced an adversarial morphing attack to evade FR systems by manipulating facial content \cite{digital_attack4}; in this attack the pixels were perturbed spatially in a coherent manner. 
\citet{digital_attack5} proposed a partial face tampering attack in which facial regions were replaced or morphed to generate adversarial samples.

All of the above mentioned studies digitally manipulated the input images of the FR models in order to fool them. In some of the studies, these manipulations were easy to detect, while in others, the manipulations were barely distinguishable; regardless, it is very difficult to transfer such digital manipulation to the real world.

\subsubsection{AML attacks in the physical domain}\label{subsubsection:Physical attacks}     

In a recent study, a printable sticker attached to a hat was used to fool the FR system; in this case, the perturbation generated was restricted to the rectangular shape of the sticker \cite{patches-attack2}. 
\citet{glasses-attack1} used printable eyeglass frames to manipulate FR models.
The perturbations were generated by gradient-based methods and were restricted to the shape of the eyeglass frame. 
Later research showed that the eyeglass frames could also be produced using generative networks \cite{glasses-attack2}.
In each of the above studies, physical objects aimed at fooling FR systems were created; the studies also share a significant disadvantage as in each case, the object created has an abnormal appearance (a low level of physical inconspicuousness). In addition, the attacks are not practical, since in many real-life situations involving automated recognition processes (e.g., at airports) such objects must be removed.

\subsection{Makeup in the Context of FR Systems}\label{subsection:Makeup works}
The application of cosmetics can substantially alter an individual's appearance and thus transform images of their face. This type of beauty enhancement can have a negative effect on the performance of FR systems \cite{makeup_survey,makeup_survey2}.
One of the first studies in this domain explored the impact of makeup on automatic FR algorithms by creating two datasets: virtual makeup (VMU) and YouTube makeup (YMU) \cite{makeup_datasets}.
The YMU dataset consists of face images of young Caucasian females obtained from YouTube makeup tutorials where the makeup was physically applied on each subject. The VMU dataset contains face images of Caucasian females where the makeup was applied digitally on each face image, using three different degrees of makeup application: lipstick, eye makeup, and full makeup.
Then in \cite{makeup-attack-physical}, another dataset, the makeup induced face spoofing (MIFS) datasaet, was used to examine makeup's spoofing capabilities on FR systems. The MIFS dataset images were collected from random YouTube makeup video tutorials in which the makeup was physically applied to each subject.
A recent study presents a new digitally makeup dataset which is purpose is to cause impersonation and identity concealment. \cite{makeup_dataset_2021}
While both first studies found that makeup has a negative effect on the performance of FR systems, they did not create makeup on the face images to attack an FR system.
The last study was created in order to attack such systems, but it focuses only on the digital domain. Moreover, in many of the these datasets' images, the makeup is heavily applied and does not look natural.

Another study suggested to transfer non-makeup images to makeup images where the perturbation information of the attack is hidden in the makeup areas~\cite{makeup-attack-digital}. 
However, in this case, the perturbation generated and added to the face images is digital noise, and such perturbation is difficult to reconstruct in the real world; in addition, the process of concealing the adversarial perturbation using natural-looking makeup poses a challenge. 
In another innovative study that used makeup to launch an attack against FR systems, the authors both digitally added makeup to the eye region of face images and considered a physical attack scenario where they attached an eye-shadow tattoo patch to the orbital region of a human subject~\cite{advmakeup2021}.
The main limitation of this study is the low physical inconspicuousness of the patch.

\section{Proposed Dodging Attack Using Makeup}\label{section:Methodology}

\subsection{Face Recognition Use Case}
Let us assume a surveillance system that is used to identify $\{p_1, p_2, ..., p_w\} \in P$ black-listed individuals, where $p_i$ represents the embedding vector of an image related to an individual $i$ on the black-list. The system receives a stream of images $\{s_1, s_2, ..., s_n\} \in S$ that were captured by a camera. A face detector model $D$ is applied on each image $s_i$, and every detected face $\{f_1,...,f_m\} \in F$ is sent to the FR model $M$. The FR model $M$ converts each $f_i$ image into an embedding vector $v_i$. Each vector $v_i$ is compared with all of the vectors in $P$. If the similarity between the vectors exceeds a certain threshold, an identification alert is raised.

\begin{figure}[t]
    \begin{center}
    \includegraphics[width=0.9\linewidth]{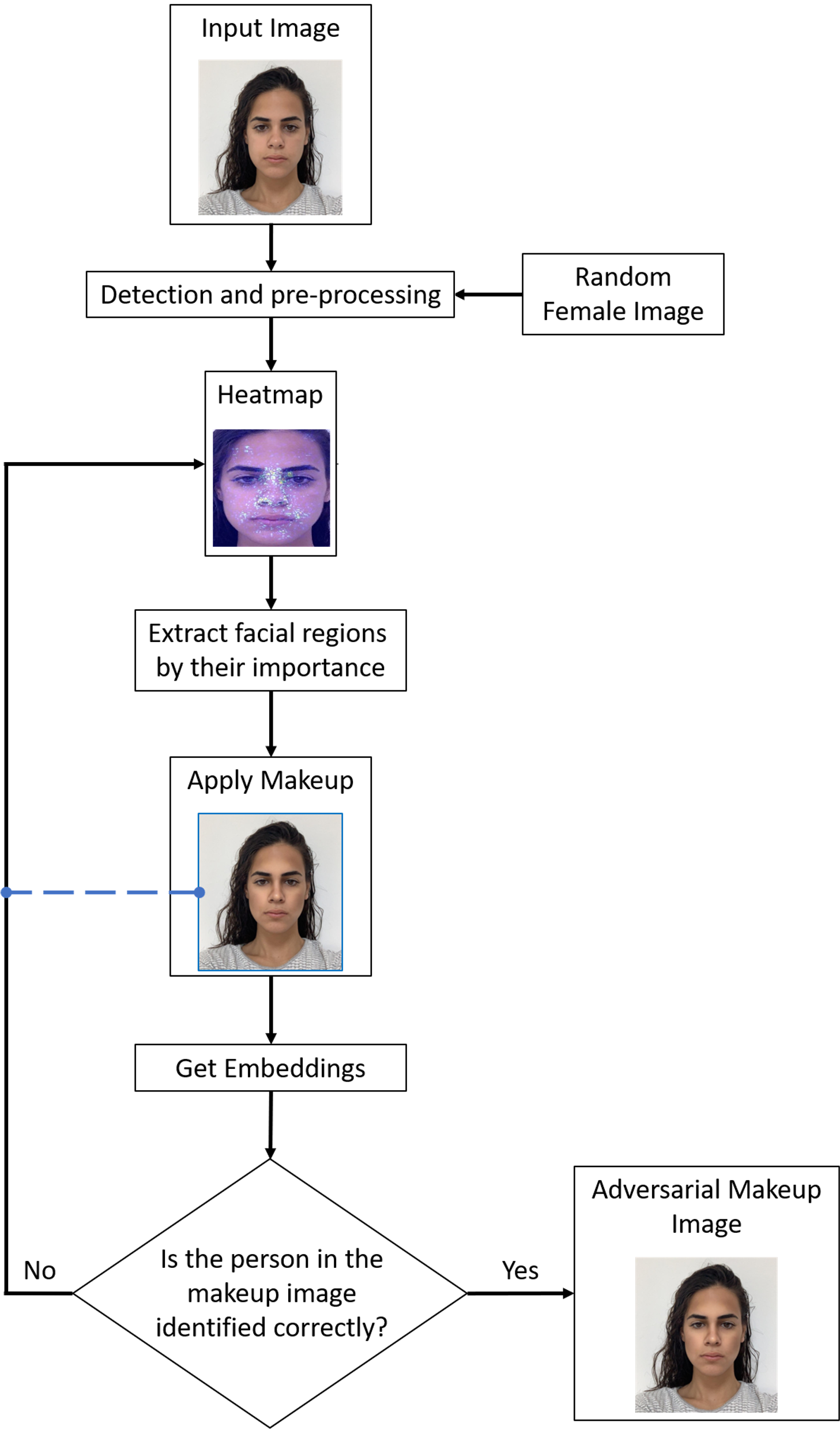}
    \end{center}
    \caption{Creation of digital adversarial makeup: a heatmap is calculated for the input image, and then makeup is added to the relevant facial regions in the image. As long as the new makeup image is still correctly recognized, it is used as input; the process is repeated until the image is no longer correctly recognized.}
\label{attack pipeline}
\end{figure}

\subsection{Threat Model}\label{subsection:Threat model}
The main goal of this study is to enable an attacker to evade identification by a FR system in the real world by applying natural-looking makeup. 
The physical makeup will be transferred from a digital makeup created by the surrogate model.

Transferability is a known property of adversarial examples  \cite{transferability1}, and given two different models that are trained on different datasets but perform the same task, adversarial examples that have been designed to fool one model often fool other models as well. FR models perform the same task of mapping face images to some vector space. By creating adversarial examples for face images using makeup elements designed to fool one FR model (a surrogate model), we presume that we can fool other target FR models with high probability. Our attacker assumes a black-box scenario, meaning that the attacker cannot access the target FR model, its architecture,  or any of its parameters. Therefore, attacker's only option is to alter his/her face before being captured by the cameras that feeds the input to the target FR model. Thus, to launch the attack, i.e., to calculate the adversarial makeup required, our attacker uses one model (the \textit{FaceNet} model \cite{FaceRecognitionNN2}) as a surrogate model, whose architecture and parameters are known. Our attacker assumes that the adversarial makeup created based on the surrogate model will fool the target model (\textit{LResNet100E-IR, ArcFace@ms1m-refine-v2}, \cite{ArcFace}).

\subsection{High-Level Description of the Attack}
The adversarial makeup attack includes two phases: an offline phase where the adversarial makeup attack is created in the digital domain, and an online phase where the attacker applies the adversarial makeup in the physical domain. The main steps of the attack are as follows:

\begin{itemize}
\item \textbf{Attack's offline preparation phase}: 
\begin{enumerate}
    \item \label{step_1} Given a surrogate FR model $M_s$, an image of the attacker $x$, additional images of the attacker $X$, and additional images of another identity $Y$, a heatmap that indicates the facial regions that have the greatest importance on the identification decision is computed.
    \item \label{step_2} Makeup elements are added digitally to $x$ according to the facial regions identified in the previous step, creating a new image $x_i$.
    \item The identity behind $x_i$ is tested using the FR model $M_s$, steps~\ref{step_1} and~\ref{step_2} are performed again while the next equation is still exists: $$1 - \frac{M_s(X) \cdot M_s(x_i)}{||M_s(X)||_2||M_s(x_i)||_2} < Threshold$$
    At the end of this step an adversarial makeup perturbation is obtained. 

\end{enumerate}

Figure~\ref{attack pipeline} presents the attack's offline preparation phase, the main purpose of which is to create adversarial makeup in the digital domain.

\item \textbf{Attack's online execution phase}: 
    \begin{enumerate}
    \item Given the digital makeup image $x_i$, the makeup is physically applied on the attacker with the assistance of a makeup artist.
    \item The attacker wearing the adversarial makeup attempts to evade the real-time FR system which is implemented with a target model $M_t$ that differs from the surrogate model $M_s$.
    \end{enumerate}
\end{itemize}

\subsection{Surrogate Model Explainability}\label{subsection:Explainability}
To create an adversarial makeup perturbation that dodges the FR system, we first need to know what areas in the face have the greatest influence on the identification decision of the FR model. Therefore, in the attack's offline preparation phase we use a heatmap to represent these areas. 
To create the heatmap, we calculate the backward gradients of a variant triplet loss function~\cite{FaceRecognitionNN2} for a tested attacker's face image using different face images of the attacker, and of an additional random non black-listed identity. Then, the result is projected on the tested face image of the attacker.
The heatmap provides a convenient visualization of the areas of the face that are important for identification (like the nose, cheeks, etc.). The importance of each facial region is represented by the intensity of the gradients, where higher intensity represents greater importance. During the attack, natural-looking makeup will be applied to the identified areas; the resulting image will then be tested against the FR system implemented with the surrogate model in order to determine whether the proposed makeup was able to hide the identity.

\noindent Our method uses a variant of the classic triplet loss function: $$L = \max(0,||f(x_i^a) - f(x_i^n)||_2^2 - ||f(x_i^a) - f(x_i^p)||_2^2 + 0.8)$$
where $L$ is used to maximize the distance between the anchor $x_i^a$ and the positive $x_i^p$ images and minimize the distance between the anchor and the negative $x_i^n$ images that we chose randomly from a dataset that contains non-black-listed identities. This way, we can influence the attacker's identity so it is closer to a non-black-listed identity in the identification process.

\section{Experiments}\label{section:Evaluation}

\subsection{Evaluation Process}
To launch the attack, we obtained frontal digital images of each black-listed participant; then following our proposed methodology, we created adversarial digital makeup images that dodge the surrogate model. These digital adversarial examples were then tested against another model that represents the target attack model. To assess the dodging attack's success in the real world under live video conditions, a makeup artist physically applied the makeup on the participant's face as it was applied to the image in the digital case. Then we used the target model to evaluate the FR system's ability to identify the participants wearing the makeup.

\subsection{Adversarial Makeup Perturbation}\label{subsection:Makeup calculation}
We created an adversarial makeup perturbation which, when added to the attacker's image, is able to dodge the surrogate model.
We only used frontal face images. 
Each image was aligned with \textit{MTCNN}~\cite{MTCNN} detector landmarks and transformed into an image of 160x160 pixels to serve as the surrogate model's input. 
We used \textit{YouCam Makeup}, a manual virtual makeup tool, to add the makeup. To obtain a natural-looking image, we only used: (1) conventional makeup techniques for using eyeshadow, eyelashes, blush, and lipstick, and contouring, and changing the shape of the eyebrows; the makeup was not used to draw any shapes (e.g., flowers, beauty marks); and (2) daily use color palettes, where dark and light brown colors were used for contouring, and only colors used in daily makeup were used for lips and eyebrows (meaning that blue, green, white, and other colors that are not applied in daily makeup were not used). 
The final adversarial example was obtained using an iterative process, whereby given an input image $x$, in each attack step $i$, makeup is applied to the facial region that has the greatest importance to the identification decision of the FR model. The creation of the adversarial makeup is finished when an image $x_i$ is not identified as a black-listed identity. Adding more makeup to an image increases the intensity of the makeup, and since we want to achieve a natural look, we stop adding makeup at this stage. During the experiments we set the threshold of the surrogate model (\textit{LResNet100E-IR, ArcFace@ms1m-refine-v2}) at 0.368.

\subsection{Testbed}
\label{sec:testbed}
\begin{figure}[t]
    \begin{center}
    \includegraphics[width=0.6\linewidth]{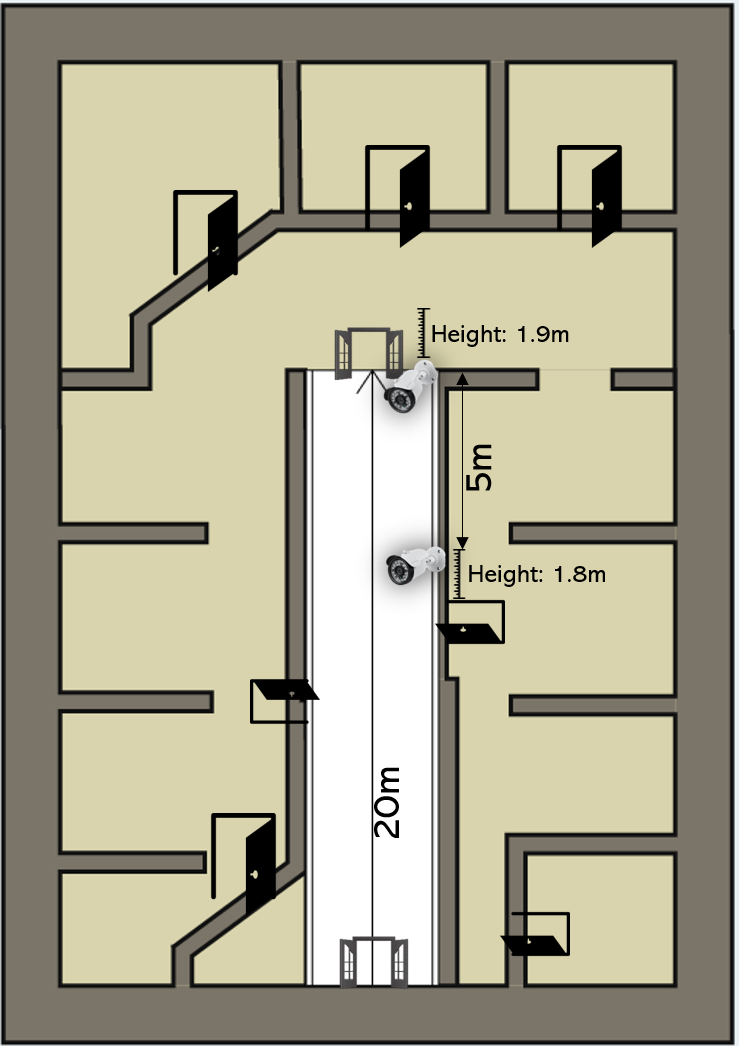}
    \end{center}
    \caption{A schematic top view of the physical evaluation setup: the white space shows the observed corridor; the position and height of each camera are noted, and the middle arrow indicates the walking direction.}
\label{fig:Evaluation setup top view}
\end{figure}

To evaluate the attack in a real-world scenario, we designed the testbed illustrated in Figure~\ref{fig:Evaluation setup top view}. 
Our testbed contains an FR system that analyzes a live video stream from two different cameras and raises an alert if a black-listed individual's face is captured.
The FR system was implemented with the following configurations:
\begin{itemize}
    \item \textbf{Physical setup}: Two cameras were installed in a corridor; each camera has the following configuration: full HD 1080p resolution (1920 x 1080), transmission in an H.265 encoding, a maximum bitrate of 1280 Kb/s, and a frame rate of 10 FPS; the brightness of the cameras was set at 60, and the mode was ``normal." 
    \item \textbf{Face detector}: The \textit{MTCNN} face detection model~\cite{MTCNN} was used with a minimum size of the detected face image to be 15x15.
    \item \textbf{Attacked (target) model}: The \textit{LResNet100E-IR, ArcFace@ms1m-refine-v2}, a pretrained FR model\footnote{\url{https://github.com/deepinsight/insightface/tree/master/model_zoo}} with an input image size of 112x112 pixels was used.
    \item \textbf{Similarity function and thresholds}: The cosine distance was used for the similarity verification function, and the verification threshold, which was calculated according to all of the LFW \cite{lfw} dataset's identities, was set at 0.42. In the physical setup, we tested the performance of the system with a live video stream; therefore, we defined a persistency threshold of seven frames before the system raises an alert that a black-listed individual has been identified. This threshold was used to minimize false positive alarms.
    
\end{itemize}

\subsection{Dataset}
Our experiments were performed on a group of 10 males and 10 females, ages 20-28, after obtaining approval from the university's ethics committee. Each participant was defined as a black-listed person in our system. We used three frontal images of each participant, two of which were enrolled in the system; the third one was used to create two additional images: one with digital adversarial makeup and the other with random makeup.
We also included all of the identities in the LFW dataset~\cite{lfw} in our system, 10 non-makeup female identities from the YMU dataset~\cite{makeup_datasets}, and an additional 10 non-makeup male identities to create the heatmaps by random choice of one of these identities for each participant. 
The digital images were tested against our FR system that was implemented with the target model. 
During the evaluation in the physical domain, each participant was requested to walk along the corridor three times: without makeup, with adversarial makeup, and with random natural-looking makeup.

\subsection{Evaluation Measurements}\label{subsection:Evaluation measurements}
\begin{figure}[t]
    \begin{center}
    \includegraphics[width=0.99\linewidth]{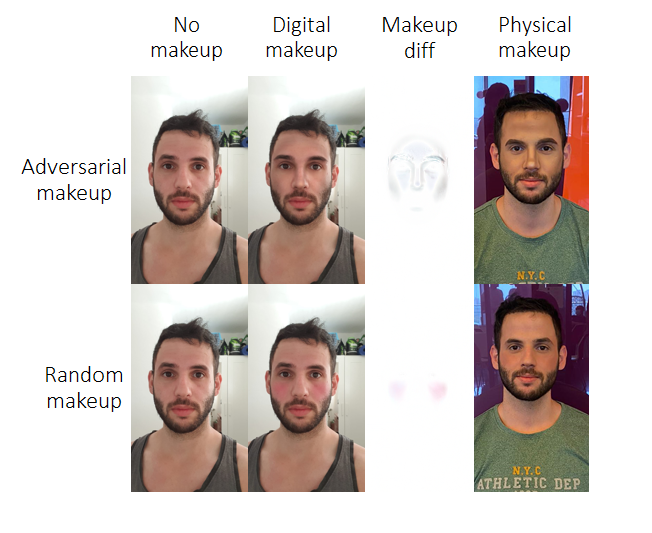}
    \end{center}
    \caption{Participant number three with adversarial makeup and random makeup. The digital images' intensity scores are 1.81 and 2.30 for the adversarial and random makeup methods, respectively.}
\label{makeup methods}
\end{figure}

We examined three different identification setups: without makeup, with adversarial makeup, and with random makeup. 
The random makeup is generated randomly such that its intensity is similar to or higher than the adversarial makeup.
An example of the random and adversarial makeup methods is presented in Figure~\ref{makeup methods}.
To measure the intensity of the digital makeup images, the adversarial makeup, and the random makeup, we utilized the colorfulness measure~\cite{colorfulness_measurment} which uses an opponent color space, with normalization of the mean and standard deviation. We applied this measure on the makeup perturbation, which is the difference between an image with makeup and an image without makeup.

For the physical attack on a live video stream, we evaluated the FR model's performance with the following measure that quantifies the recognition rate: $$r_{rec} = \frac{|R_i|}{|R_i|+|D_i|}$$
where $R_i$ is the number of frames in which the participant $i$ was recognized, and $D_i$ is the number of frames in which the participant $i$ was detected but not recognized.

\subsection{Results}\label{subsection:Results}
\subsubsection{Evaluation in the digital domain}
In this evaluation, we only examined frontal face images of our black-listed participants. We assessed the influence of the random natural-looking digital makeup and the adversarial natural digital makeup on the FR system's performance. First, we verified that the original images were identified correctly by the target model. Then, we added makeup to these images using two methods (random and adversarial), and we measured the model's recognition rate.

When examining the digital adversarial makeup images we found that not all of the individuals were identified as their ground truth identity by both models, whereas all of the random makeup images were correctly identified. 
These results may indicate that the use of non-adversarial (random) makeup will not fool an FR system.
The average intensity score of the random makeup images is 2.85, while that of the adversarial makeup images is 2.64. This indicates that our adversarial makeup's intensity is lower than that of random makeup, while being more effective at dodging the FR system. 

\subsubsection{Evaluation in the physical domain}
In this evaluation, we examined the physical makeup's influence on identification decision of the FR system using live video streams of the black-listed participants.
Similar to the digital domain, we also verified that each participant without makeup was identified correctly by the FR system in real-time. Then, each participant's identity was evaluated again with the two makeup methods: random and adversarial. A makeup artist used the digital makeup images to physically apply natural-looking makeup on the human participants.
The performance of our adversarial makeup attack was evaluated using the FR system, and the recognition rate was computed using three video streams for each participant (without makeup, with random makeup, and with adversarial makeup). 

Table~\ref{table:results} presents the evaluation results for each participant, and the average results are presented in Table~\ref{table:avg_results}. 
The results presented in Table~\ref{table:avg_results} show that the random natural-looking makeup decreases the recognition rate slightly from 42.61\% (female) and 52.53\% (male) to 31.69\% (female) and 35.78\% (male). The use of the adversarial natural-looking makeup decreases the recognition rate dramatically from 42.61\% (female) and 52.53\% (male) to 0.9\% (female) and 1.53\% (male), which is below a reasonable threshold of a realistic operational environment.  

\begin{table}[t!]
\caption{Physical attack's average recognition rate} 
\centering 
\begin{tabular}{@{}*{4}{c}@{}} 
\hline 
& \multicolumn{3}{c}{Makeup}\\
&{no} & {random} & {adversarial}\\ [0.5ex]
\hline 

Females & 42.61\%	& 31.69\% & 0.90\%\\[1ex]
Males & 52.53\% & 35.78\% & 1.53\% \\[1ex]
Both & 47.57\%& 33.73\% & 1.22\% \\[1ex]

\hline 
\end{tabular}
\label{table:avg_results}
\end{table}

\section{Conclusions}\label{section:Conclusions}
In this study, we showed how natural-looking makeup can be used to prevent an FR system from identifying black-listed individuals. In contrast to methods proposed in prior research, our attack has a high level of inconspicuousness and thus can be launched in the real world. We demonstrated the effectiveness of our attack under black-box conditions, using a testbed with two cameras that mimics a realistic real-time survailance system. The evaluation results show that in the digital domain, the face recognition system was unable to identify all of the participants, while in a physical domain, the face recognition system was only able to identify the participants in 1.22\% of the frames (compared to 47.57\% without makeup and 33.73\% with random natural-looking makeup), which is below a reasonable threshold of a realistic operational environment.

\begin{table}[H]
\caption{Real-time physical makeup attack results} 
\centering 
\scriptsize
\begin{tabular} {|c@{\hspace{1\tabcolsep}} c@{\hspace{1\tabcolsep}} c@{\hspace{1\tabcolsep}}| c@{\hspace{1\tabcolsep}} c@{\hspace{1\tabcolsep}} c@{\hspace{1\tabcolsep}}| c@{\hspace{1\tabcolsep}} c@{\hspace{1\tabcolsep}}|}

\hline
ID & {Gender} & {Method} & {All Cam} & {Cam \#1} & {Cam \#2} & {Cam \#1} & {Cam \#2}
\\ [0.5ex]

 & & & \multicolumn{3}{c|}{\textit{recognition rate}} & \multicolumn{2}{c|}{$alarm$}  \\
\hline
&&no &31.33\% & 25.53\% & 38.89\% & T & T \\[0ex]
1 &F & rnd & 11.63\% & 3.0\% & 23.61\%  & T & T \\[0ex]
& &adv &0.0\% & 0.0\% & 0.0\%  & F & F \\[0.3ex]

& &no &43.4\% & 42.86\% & 43.86\%  & T & T \\[0ex]
2 &F &rnd & 28.99\% & 32.35\% & 25.71\%  & T & T \\[0ex]
& &adv &1.37\% & 3.03\% & 0.0\%  & F & F \\[0.3ex]

& &no &49.52\% & 55.77\% & 43.4\%  & T & T \\[0ex]
3 &M &rnd & 24.8\% & 28.81\% & 21.21\%  & T & T \\[0ex]
& &adv &0.0\% & 0.0\% & 0.0\%  & F & F \\[0.3ex]

& &no &61.95\% & 46.15\% & 75.41\% & T & T \\[0ex]
4 &M &rnd & 42.5\% & 30.36\% & 53.12\% & T & T \\[0ex]
& &adv &0.88\% & 0.0\% & 1.64\% & F & F \\[0.3ex]

& &no &32.67\% & 25.0\% & 39.62\% & T & T \\[0ex]
5&F &rnd & 12.96\% & 7.84\% & 17.54\% & T & T \\[0ex]
& &adv &0.0\% & 0.0\% & 0.0\% & F & F  \\[0.3ex]

& &no &66.07\% & 66.0\% & 66.13\% & T & T \\[0ex]
6&M &rnd & 47.33\% & 50.82\% & 44.29\% & T & T \\[0ex]
& &adv &0.0\% & 0.0\% & 0.0\% & F & F \\[0.3ex]

& &no &68.14\% & 64.81\% & 71.19\% & T & T \\[0ex]
7&M &rnd & 25.96\% & 17.65\% & 33.96\% & T & T \\[0ex]
& &adv &3.88\% & 0.0\% & 7.41\% & F & F \\[0.3ex]

& &no &52.13\% & 59.46\% & 47.37\% & T & T \\[0ex]
8&M &rnd & 31.3\% & 36.54\% & 26.98\% & T & T \\[0ex]
& &adv &0.0\% & 0.0\% & 0.0\% & F & F \\[0.3ex]

& &no &34.74\% & 33.33\% & 36.0\% & T & T \\[0ex]
9&F &rnd & 4.39\% & 8.16\% & 1.54\% & F & F \\[0ex]
& &adv &0.0\% & 0.0\% & 0.0\% & F & F \\[0.3ex]

& &no &27.47\% & 32.56\% & 22.92\% & T & T \\[0ex]
10&M &rnd & 33.06\% & 19.61\% & 42.47\% & T & T \\[0ex]
& &adv &2.94\% & 6.52\% & 0.0\% & F & F \\[0.3ex]

& &no &61.64\% & 53.23\% & 67.86\% & T & T \\[0ex]
11&M &rnd & 34.13\% & 29.82\% & 37.68\% & T & T \\[0ex]
& &adv &5.22\% & 0.0\% & 8.7\% & F & F \\[0.3ex]

& &no &49.52\% & 60.78\% & 38.89\% & T & T \\[0ex]
12&F &rnd& 54.46\% & 41.67\% & 66.04\% & T & T \\[0ex]
& &adv &5.1\% & 2.38\% & 7.14\% & F & F \\[0.3ex]

& &no &30.61\% & 29.41\% & 31.91\% & T & T \\[0ex]
13&M &rnd & 17.78\% & 10.53\% & 23.08\% & T & T \\[0ex]
& &adv &0.0\% & 0.0\% & 0.0\% & F & F \\[0.3ex]

& &no &66.2\% & 65.62\% & 66.67\% & T & T \\[0ex]
14&M &rnd & 65.0\% & 64.86\% & 65.12\% & T & T \\[0ex]
& &adv &2.41\% & 0.0\% & 4.76\% & F & F \\[0.3ex]

& &no &40.2\% & 48.89\% & 33.33\% & T & T \\[0ex]
15&F &rnd& 39.06\% & 33.33\% & 42.5\% & T & T \\[0ex]
& &adv &0.0\% & 0.0\% & 0.0\% & F & F \\[0.3ex]

 & &no &33.76\% & 42.11\% & 25.93\% & T & T \\[0ex]
16& F &rnd & 34.25\% & 44.09\% & 23.86\% & T & T \\[0ex]
& &adv &0.0\% & 0.0\% & 0.0\% & F & F \\[0.3ex]

& &no &73.68\% & 66.13\% & 80.28\% & T & T \\[0ex]
17& F &rnd & 64.29\% & 57.89\% & 70.91\% & T & T \\[0ex]
& &adv &2.5\% & 5.13\% & 0.0\% & F & F \\[0.3ex]

& &no &20.69\% & 18.75\% & 22.22\% & T & T \\[0ex]
18& F &rnd & 11.52\% & 12.36\% & 10.78\% & T & T \\[0ex]
& &adv &0.0\% & 0.0\% & 0.0\% & F & F \\[0.3ex]

 & &no &66.1\% & 64.56\% & 67.35\% & T & T \\[0ex]
19& F &rnd & 55.3\% & 46.77\% & 62.86\% & T & T \\[0ex]
& &adv &0.0\% & 0.0\% & 0.0\% & F & F \\[0.3ex]

 & &no &41.59\% & 44.23\% & 39.34\% & T & T \\[0ex]
20& M &rnd & 35.96\% & 44.23\% & 29.03\% & T & T \\[0ex]
&&adv &0.0\% & 0.0\% & 0.0\% & F & F \\[0.3ex]

\hline 
\end{tabular}
\label{table:results}
\end{table}

\bibliography{ref}



\end{document}